\documentclass[12pt]{iopart}
\usepackage{cite}
\usepackage{graphicx}
\usepackage{multirow}
\usepackage{amssymb}
\usepackage{xcolor}
\usepackage{orcidlink}
\usepackage{fontawesome5}
\usepackage[utf8]{inputenc}

\begin{document}

\title[]{Lightweight wood panel defect detection method incorporating attention mechanism and feature fusion network}

\author{Yongxin Cao, Fanghua Liu, Lai Jiang, Cheng Bao, You Miao and Yang Chen}



\begin{abstract}
In recent years, deep learning has made significant progress in wood panel defect detection. However, there are still challenges such as low detection , slow detection speed, and difficulties in deploying embedded devices on wood panel surfaces. To overcome these issues, we propose a lightweight wood panel defect detection method called YOLOv5-LW, which incorporates attention mechanisms and a feature fusion network.
Firstly, to enhance the detection capability of acceptable defects, we introduce the Multi-scale Bi-directional Feature Pyramid Network (MBiFPN) as a feature fusion network. The MBiFPN reduces feature loss, enriches local and detailed features, and improves the model's detection capability for acceptable defects. 
Secondly, to achieve a lightweight design, we reconstruct the ShuffleNetv2 network model as the backbone network. This reconstruction reduces the number of parameters and computational requirements while maintaining performance. We also introduce the Stem Block and Spatial Pyramid Pooling Fast (SPPF) models to compensate for any accuracy loss resulting from the lightweight design, ensuring the model's detection capabilities remain intact while being computationally efficient.
Thirdly, we enhance the backbone network by incorporating Efficient Channel Attention (ECA), which improves the network's focus on key information relevant to defect detection. By attending to essential features, the model becomes more proficient in accurately identifying and localizing defects.
We validate the proposed method using a self-developed wood panel defect dataset. 
The experimental results demonstrate the effectiveness of the improved YOLOv5-LW method. Compared to the original model, our approach achieves a 92.8\% accuracy rate, reduces the number of parameters by 27.78\%, compresses computational volume by 41.25\%, improves detection inference speed by 10.16\%, and enhances the detection accuracy of two types of acceptable defects (dead knots and cracks) by 0.2\% and 1.3\% respectively.

\end{abstract}
%
\vspace{1pc}
\noindent{\it Keywords}: wood panel defect detection, YOLOv5, lightweight, feature fusion, attention mechanism
%
%
\maketitle
%
%

\section{Introduction}

Wood panel defect detection \cite{liu2022low, xie2023detection, zhang2022intelligent, wang2022bolt, Song_2023} is very important in this day and age. Especially in the decoration industry \cite{aktas2012impact}, wood plays a direct role as a very important raw material and its quality is good or bad. In addition to this, consumers prefer to purchase solid wood materials that are free of knots, cracks, and holes. To meet the needs of consumers for wood panels, wood processing companies have to spend a lot of labour to identify the types of surface defects in the wood, remove the defective pieces and splice the remaining material together, a popular method among wood processing companies, which not only reduces the waste of wood but also increases the company's profitability. However, the use of large numbers of people to identify surface defects in timber can have the disadvantages of high labour costs, low efficiency, and susceptibility to subjective judgment \cite{khan2022optimization}.

Traditional methods of detecting wood defects include ultrasonic detection \cite{zhao2017quantitative, peng2016simultaneous}, laser detection \cite{fang2017review, wang2009pattern}, X-ray detection \cite{krahenbuhl2014knot, sarigul2003rule}, and infrared detection \cite{quin1998locating}. However, the cost of their equipment is high, and the speed of detection is also slow, making them less widely applicable. In recent years, machine vision-based image inspection technology has developed rapidly. The traditional machine vision method of recognizing board defects is to identify board defects after steps such as greyscale transformation, smoothing filtering, threshold segmentation, edge detection, and contour extraction \cite{ruz2009automated, yang2018wood, zhang2016identification}. The accuracy of this recognition method is low, stability is poor, and the detection speed does not meet the real-time requirements. With the development of traditional target detection algorithms, performance improvements have also encountered bottlenecks \cite{zou2023object}. Mainly, the following drawbacks exist: a large number of redundant regions are generated along with candidate regions; it is difficult to extract semantic information-rich regions from complex images \cite{wu2020recent}. Therefore, for some scenes with complex backgrounds, uneven illumination, and small defects, traditional target detection algorithms \cite{cheng2022moving} suffer from missed and false detection, resulting in low detection accuracy, poor real-time performance, and weak generalization ability.

\begin{figure}[t]
\centering
\includegraphics[width=1\textwidth]{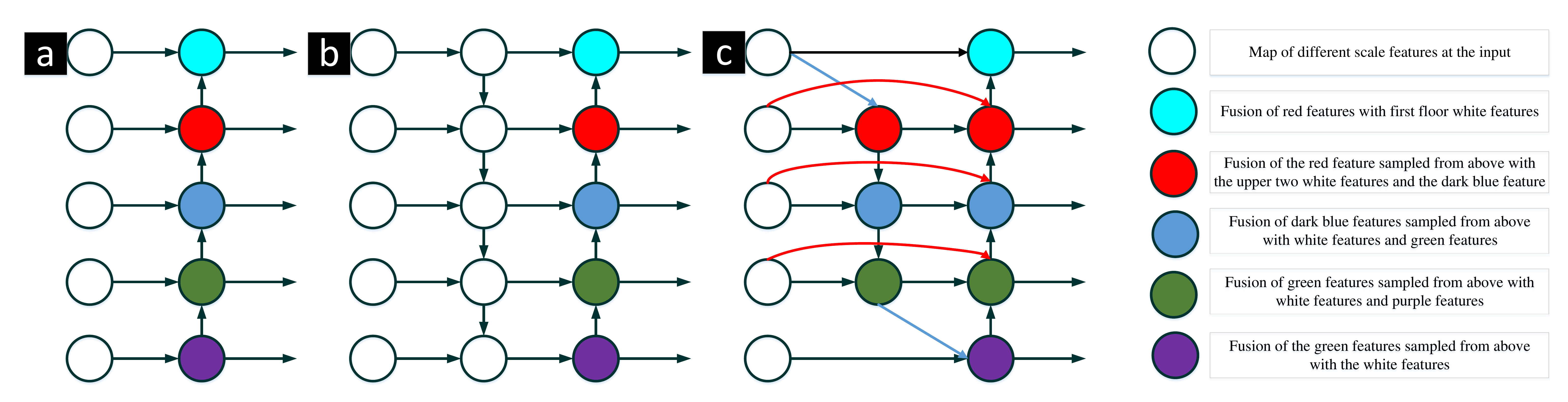}
\caption{Feature Fusion Network: (a) FPN; (b) PANet; (c) MBiFPN. The MBiFPN feature fusion network eliminates nodes with only one input on top of the PANet to speed up feature fusion efficiency, and adds cross-layer connections at the same layer to allow features to be fused more fully, with each node having two inputs and one output.\label{fig1}}
\end{figure}

In recent years, many methods \cite{wang2022improved, Liu_2020, wang2021channel} have applied deep learning to industry, most notably because deep learning has great advantages in dealing with high-dimensional data. It uses end-to-end layer-by-layer learning to mine and learn data features from shallow to deep layers, which is suitable for large amounts of data. One of the most researched boomers is convolutional neural networks. Wu et al \cite{wu2020using} used YOLOv4 deep learning network in agriculture to detect apple blossoms and achieved good results. Song et al \cite{Song_2023} improved YOLOv4 for spatial positioning of car door frames and achieved a positioning error of less than 1cm. In addition, it is also widely used in the field of wood panelling. He et al \cite{he2020application} improved DCNN deep convolutional neural network training with data augmented dataset and obtained a high accuracy rate, identifying wood defects more effectively and accurately. However, an easy-to-deploy network is yet to be developed for wood panel defect detection.

To solve the problem that the current network model is difficult to deploy with a large number of parameters, slow in recognition, and difficult to adapt to the detection of fine wood panel cracks and dead knot defects, this paper proposes a lightweight wood panel defect detection method YOLOv5-LW based on YOLOv5. Firstly, the $S^3$Net (ShuffleNetV2+Stem Block+SPPF) method is used to replace the backbone network for sampling to maintain certain accuracy while reducing the model parameters, and secondly, the Efficient Channel Attention (ECA) module is added after the feature maps are extracted by the backbone network to improve the network's ability to detect fine cracks while being as lightweight as possible. Finally, the neck section uses the Multi-scale Bi-directional Feature Pyramid Network (MBiFPN) to replace the FPN structure to achieve multi-scale feature fusion and improve the detection accuracy of the model. As can be seen from Figure~\ref{fig1}, Feature Pyramid Networks (FPN) feature fusion is a top-down fusion approach with a single path, which also results in information loss. In contrast, the Path Aggregation Network (PANet) feature fusion network not only fuses top-down but also adds an additional bottom-up fusion, which makes the underlying feature details fused more fully and the feature information transfer less efficient. The MBiFPN feature fusion network proposed in this paper eliminates some feature layer points with only one input node, fuses the small-scale feature map with the upper layer features to speed up the network express efficiency, and additionally adds a cross-layer connection from the original input to the output port, so that the nodes of the whole network can keep two inputs and one output, and the feature details are better fused to improve the defective target perception.

The contributions of this work can be summarized as follows:

(1) Enhanced Lightweight Model: We enhance a lightweight fine wood panel defect detection model by combining YOLOv5, fused attention mechanisms, and a feature fusion network. This approach allows us to achieve a balance between detection accuracy and computational efficiency, making it suitable for real-time applications.

(2) Dataset: We have curated a unique wood panel defect dataset comprising four types of defects commonly found on the surface of wood panel timber: dead knots, live knots, knot with crack, and cracks. This dataset serves as a valuable resource for training and evaluating the performance of the proposed model.

(3) Performance Evaluation: We conduct extensive experiments to evaluate the performance of the improved model for fine wood panel defect detection. In addition, we compare its performance against other classical models to provide a comprehensive analysis of its advantages and disadvantages in terms of detection accuracy, computational requirements, and speed.

\section{Related research}
\subsection{Machine vision-based target detection}
As computer intelligence recognition algorithms \cite{shen2023triplet} are widely used in various fields, research scholars from various countries have conducted relevant studies for the recognition of defects on the surface of wood panels in response to the human resource cost problem. In 2003 JW Funck \cite{funck2003image} introduced various image segmentation algorithms for subtle defects in wood panels and proposed a region-based similarity algorithm combining clustering and region-growing techniques to show the best overall performance. For defects such as image noise and low contrast, Qi et al \cite{qi2008study} proposed a BP neural network for fast and accurate identification of defects inside the template, analysed the results of different network structures and network parameters affecting the classification of wood defects, and proposed optimal network parameters for identifying the types of wood panel defects. In addition to this, Ruz et al \cite{ruz2009automated} used a fuzzy min-max neural network for image segmentation FMMIS to generate a minimum bounding rectangle for each defect present in the wood panel image with a correct classification rate of 91\%. For knot and crack defects in wood panels, in 2015 Mohamad et al \cite{hittawe2015multiple} proposed a dictionary based on SURF and LBP features and a bag-of-words approach for classification using SVM and showed through experimental results on two different datasets that the combination of the two features outperformed the single feature with a detection accuracy of 91.5\%. For wood panel accuracy improvement, Michal S et al \cite{packianather2015wrapper} analysed a method to improve the classifier by selecting the most suitable features from a given feature set, training and testing to obtain better results overall and proposed a wrapper-based feature selection method using the honeybee (Bees) algorithm for wood defect classification, with experimental results showing that a 17\% improvement. He et al \cite{he2020application} obtained 99.13\% accuracy by training a data-enhanced dataset with an improved DCNN deep convolutional neural network, which resulted in more effective and accurate identification of wood defects compared to traditional wood image recognition methods.

The YOLO algorithm is a single-stage target detection algorithm, which is seen in most industrial scenarios due to its good overall performance in terms of speed and accuracy. YOLOv5 is a single junction target detection algorithm with high detection accuracy, fast detection speed, small weights, and easy training and is suitable for embedding into devices for real-time detection. Cui et al \cite{cui2023real} proposed a basic framework based on YOLOv3, a deep learning method to improve the spatial pyramid pool SPP network, applied to the feature pyramid FP network in YOLOv3, and the experimental results showed that the overall detection accuracy reached 93.23\% and the detection time per photo was around 13ms.

\subsection{Lightweight model-based target detection}
Due to the many complexities in industrial scenarios, jitter in assembly lines, shifting positions of wood panelling, light brightness and other effects. To improve the accuracy of target recognition, deep learning models are being deepened to adapt to special complex task scenarios. But the deepening of the network inevitably leads to an increase in the number of model parameters and computations, and an increasing demand on the hardware, which is not conducive to the deployment of embedded devices. Therefore, the study of lightweight models can provide a reliable basis for the deployment of embedded devices.

Currently, some lightweight networks introduce group convolution and depth-separable convolution modules to reduce the number of convolutions in the network, increase the width of the network, and reduce network fragmentation. Some of the better-performing lightweight networks mainly include SqueezeNet, MobileNet series, ShuffleNet series, and GhostNet. Among them, the ShuffleNetv2 model was proposed by Ma et al \cite{shen2023pbsl}, which introduces point-by-point group convolution and channel mixing operation, point-by-point group convolution calculation can reduce the computation of the model, channel mixing The channel mashup solves the information interaction problem caused by group convolution. The feature channel input is divided into two branches, one of which is equivalent to the residual connection, and the other consists of three convolution modules with the same input and output. And one of the channel mixing washes disrupts the channel order of the feature map to enhance the information interaction between the two branches. Chen et al \cite{chen2022visual} used a lightweight convolutional neural network MobileNetv2 for wood quality detection to achieve efficient recognition of wood quality, improving the recognition efficiency and solving the problem of difficult applications with limited computing power and memory, but there is still the problem of difficult recognition of fine targets. problem. Wu et al \cite{wu2020using} proposed a channel pruning-based YOLOv4 model algorithm to detect apple blossoms in real-time, and the detection accuracy reached 97.31\%. wang and He et al \cite{wang2021channel} improved the light weight of YOLOv5 through pruning operation to achieve the detection of young apple fruits, and the model size was only 1.4 MB.

In summary, most of the existing algorithms for recognising surface defects in wood panels are based on traditional learning or machine learning algorithms and the recognition rate is not very high. In response to this problem, deep convolutional neural networks have become a boom today and the accuracy of defect recognition continues to improve, but the detection speed is relatively slow and does not meet the needs of the work in wood processing plants. Therefore, the following problems still need to be addressed in the task of detecting defects in wood panel timber:

(1) Defect data sets for wood panel timber, which is relatively sparse and relatively difficult to collect.

(2) A large number of network model parameters nowadays are computationally intensive and require the high performance of hardware devices, which is not conducive to the deployment of embedded devices.

(3) The detection of defects such as fine dead knots and cracks in the field of wood panel defect detection is more difficult and the recognition rate is not high.

\section{Methods}

\begin{figure}[b]
\centering
\includegraphics[width=1\textwidth,height=0.7\textwidth]{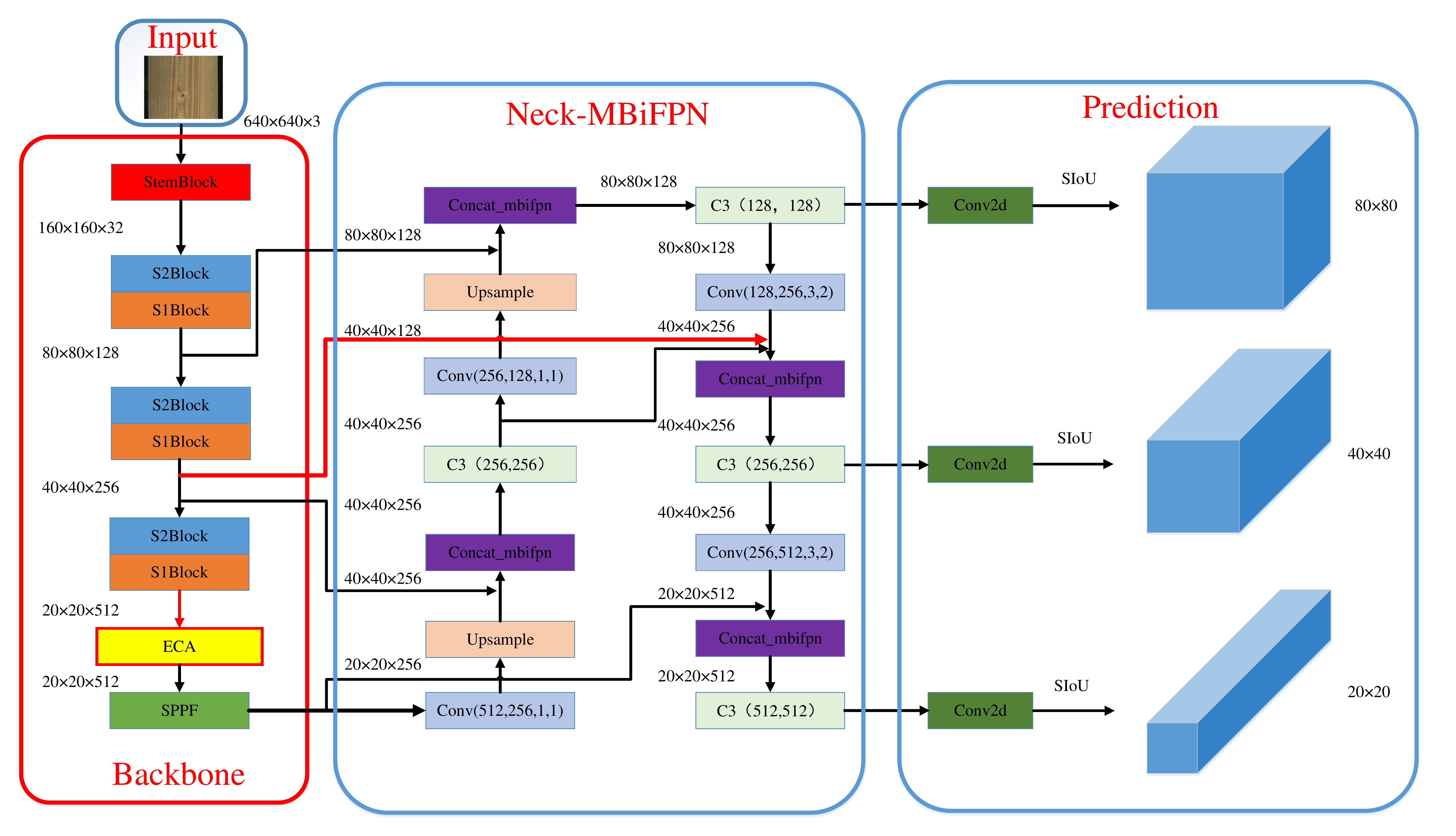}
\caption{YOLOv5-LW network method structure: the $S^3$Net is the backbone and the MBiFPN bidirectional feature pyramid network module forms the Neck part of the network model as well as the prediction head.\label{fig2}}
\end{figure}

According to the current practical application requirements of wood panel defects, this paper proposes a lightweight network YOLOv5-LW, which contains Multi-scale Bi-directional Feature Pyramid Network (MBiFPN), $S^3$Net, Efficient Channel Attention (ECA). MBiFPN constitutes the Neck part of the network model, $S^3$Net is used as the backbone network of the prototype YOLOv5 \cite{wang2022improved}, the ECA module is embedded at the bottom of the backbone network, and the network structure is shown in Figure~\ref{fig2}.

\subsection{MBiFPN in Neck}
The Feature Pyramid Networks (FPN) structure in the YOLOv5 network is a top-down channel fusion model to achieve image semantic and feature-level fusion, but the structure has the inherent limitation of unidirectional information flow. To address this issue, Path Aggregation Network (PANet) adds a bottom-up path aggregation network that can pass feature detail information from the bottom layer to the prediction head of the network, but PANet \cite{liu2018path} increases the complexity of the network and is less efficient at passing information.

\begin{figure}[b]
\centering
\includegraphics[width=1\textwidth]{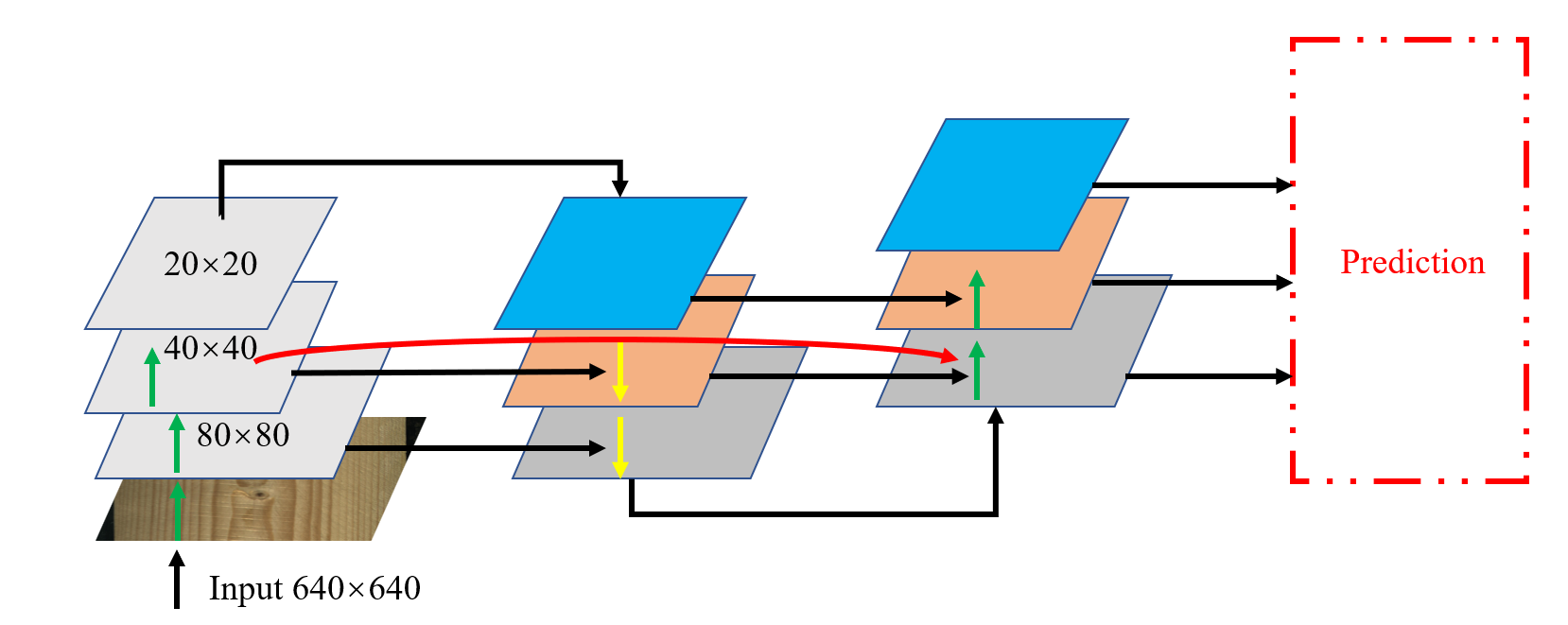}
\caption{MBiFPN introduces a 40×40 cross-layer connection and bi-directional flow feature fusion mechanism to solve the problem of information loss and redundancy in the feature pyramid network, improving the expressiveness and performance of the model.\label{fig3}}
\end{figure}

Therefore, in this paper, feature fusion is improved to Multi-scale Bi-directional Feature Pyramid Network (MBiFPN), as shown in Figure~\ref{fig3}. MBiFPN improves on PANet by removing nodes with only one input edge, which have only input edges without feature fusion and will contribute less to the network, and the removal can simplify the network. Afterwards, a new channel is added to the input node at the same scale to connect to the output node, which not only does not require a large increase in computational cost but also fuses more features. the MBiFPN architecture normalises the ratio of the weights to the sum of the weights to quickly normalise the fusion, eventually normalising it to the [0,1] interval, where information from different feature layers can be fused in the prediction layer to improve the perception of target defects in different situations.

In this paper, the 40×40 feature map obtained by 16-fold downsampling is connected to the later feature map across layers to reduce information loss and improve the network's ability to recognize small targets. Fast normalisation equation \ref{con:1} is used to keep the size of each weight between 0 and 1, which has a similarity with softmax-based fusion, but its computing performance on the hardware GPU is mentioned to be improved.

\begin{equation}
\mathrm{O}=\sum_{i=0} \frac{w_{i}}{\sum_{j=0} w_{j}+\varepsilon} \times I_{i}\label{con:1}
\end{equation}
Which, $w_i$ is the learnable weight, $I_i$  denotes the feature layer of the input.

\subsection{ShuffleNetV2+Stem Block+SPPF }
To lighten the network model, speed up the model calculation and meet the demand of model deployment of embedded devices, the $S^3$Net (ShuffleNetV2+Stem Block+SPPF) is proposed for backbone network improvement based on YOLOv5.

\begin{figure}[t]
\centering
\includegraphics[width=1\textwidth,height=0.7\textwidth]{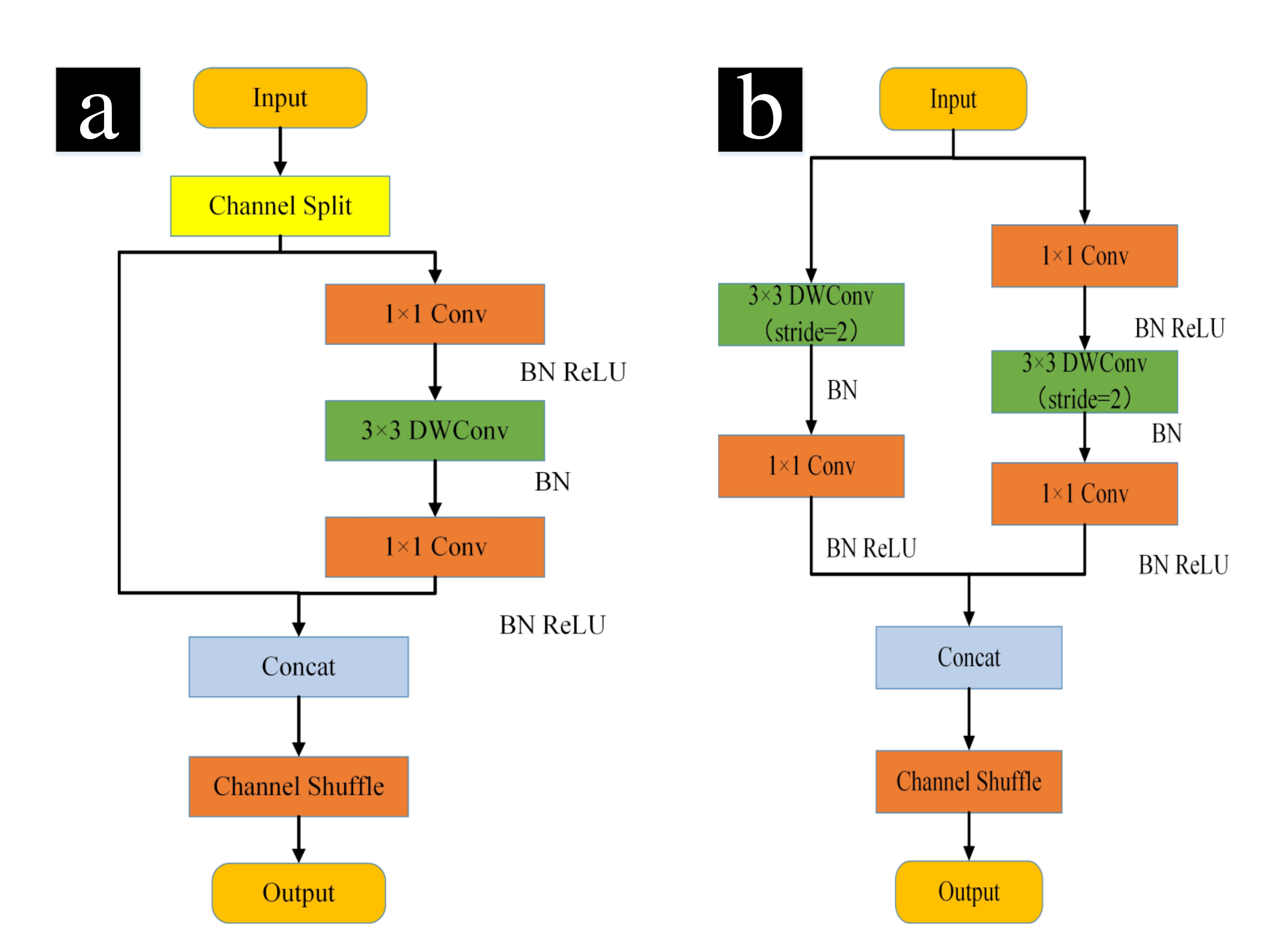}
\caption{ShufflenetV2 module: (a) S1 module; (b) S2 module. S1 module divides the number of channels of the feature map into two branches through the Concat+Channel Shuffle operation and merges them through the Channel Shuffle layer to mix and wash the features, so that the model outputs the feature map from multiple channels of the original feature map; S2 module directly inputs the feature map without using the Channel Split module to split it, and both branches are downsampled using a deep convolution with a step size of 2.\label{fig4}}
\end{figure}

In the original YOLOv5 backbone model, the network structure is mainly composed of C3 modules and Conv, with a large number of convolutional multiplication and addition operations, resulting in a large number of parameters and computational effort. In response to the need for lightweight, ShuffleNetV2 \cite{shen2023pbsl} is an improved upgrade on the version of ShuffleNetV1, which is more accurate than ShuffleNetV1 and MobileNetV2 under the same circumstances. The ShuffleNetV2 model is composed of two modules, S1 and S2, as shown in Figure~\ref{fig4} (a), with the S1 module after the feature map input by Channel Branch (Channel Split) operation, the number of feature map channels into c and c-c', the left branch to do constant mapping, increasing the depth of the network, parallel structure, speed up the model training, improve the training effect, the right branch contains three convolution operations, two of which are ordinary convolution, and finally, the two branches The S2 module Both branches are downsampled using a deep convolution with a step size of 2, and the channels are adjusted by 1×1 convolution. After convolution, the Concat+Channel Shuffle operation is used to merge. The final output of the model is a feature map with half the resolution and double the number of channels.

The first layer of the original network is composed of CBL (Conv+BN+Leaky relu) modules, which are inadequate for extracting image features and are computationally intensive. In this paper, through the method Stem Block used as a downsampling method in PeleeNet \cite{wang2018pelee}, the input feature map is first subjected to a 3×3 convolution operation to change the number of channels of the feature map, after which the feature map is also divided into two parts, one part is subjected to maximum pooling, and the other part is subjected to a 3×3 convolution with a step size of two to achieve second downsampling. The output of the two branches is then stitched together by this dimension on the channel, and finally, the number of feature map channels is reduced by 1×1 convolution. Compared with the original CBL operation, the Stem Block module allows the model to extract more feature information while reducing the number of model parameters, compensating for the loss of feature extraction due to network lightweight. In this paper, the combination of Stem Block and ShuffleNetV2 is used as the backbone part of the model and experiments are used to verify the effectiveness of the model. 
\begin{figure}[t]
\centering
\includegraphics[width=1\textwidth]{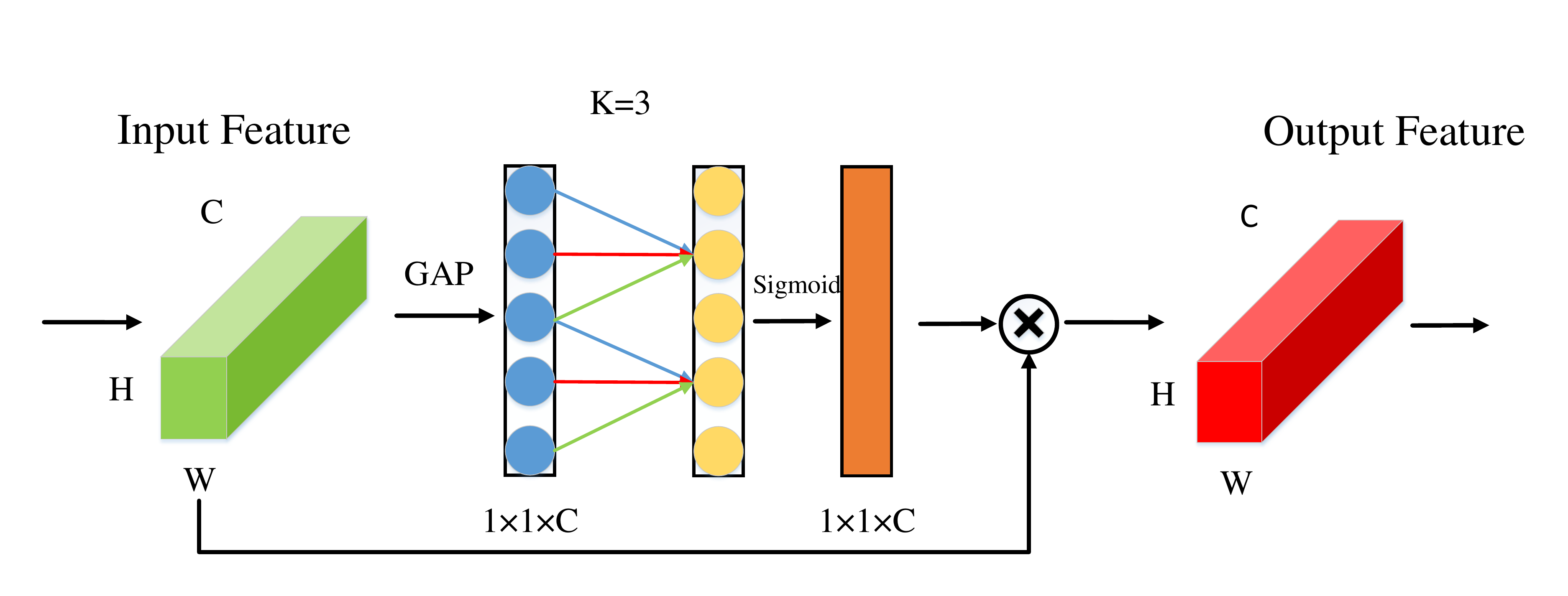}
\caption{The ECA attention mechanism calculates the weights by computing the relationship between the different positions within each channel, using a learnable scaling factor to adjust the weights, thus avoiding the computational effort of global average pooling.\label{fig5}}
\end{figure}

\subsection{ECA in Backbone}
Information attention mechanisms can be of great help in improving the performance aspects of Deep Convolutional Neural Network (DCNN). Squeeze and Excitation Network (SENet) \cite{hu2018squeeze} is a representative module that introduces channel attention to convolutional blocks and learns the channel attention of each convolutional block, resulting in a significant improvement in DCNN structure. Given the input features, the SE module first performs global level pooling for each channel individually, followed by two completed connected layers with non-linearity, and finally uses a Sigmoid function to generate the channel weights. The Efficient Channel Attention (ECA) addresses these issues by proposing a local cross-channel interaction measure that avoids dimensionality reduction through an adaptive selection of the one-dimensional convolutional kernel size. The ECA is implemented through a one-dimensional convolution of size K, where K is the coverage of the local cross-channel interaction, avoiding manual adjustment of K values and choosing an adaptive approach, where the coverage of cross-channel interactions is multiplied by the channel dimension, as shown in Figure~\ref{fig5}. The ECA module is an improvement on the SENet module, a measure that allows channels and weights to be linked while appropriate cross-channel interactions reduce the complexity of the model and maintain its performance. the ECA module is a lightweight presence in the attention mechanism and yields a significant performance increase.

\section{Self-made wood panelling data set and preparation}
Due to the lack of a large-scale database in this area of wood panel surface defects and the authenticity of the data, this paper uses the publicly available high-resolution wood surface defect images by Bodzas et al \cite{kodytek2021large}, from which four defect types, dead knots, live knots, cracks, and cracked knots, were selected, totalling 2098 images, each with a resolution of 2800×1024. Due to the small number of samples, it is necessary to expand and annotate the existing data to avoid over-fitting the model.
\subsection{Data enhancement}
In the actual working scenario of a factory, there are inevitably problems such as assembly line jitter, light colour brightness, camera angle and other changes, and therefore different patterns are presented in the images. To make the dataset closer to the real scene and meet the situation in different states, as well as to solve the problem of insufficient sample data and improve the robustness of the model, this paper performs data enhancement on the original dataset. Random methods of probabilistic enhancement such as horizontal mirror flip, changing contrast and brightness are used to increase the diversity of the data. The total number of enhanced data is 5098.
\subsection{Automatic data annotation}
As each image has one or more defective targets, the manual labelling workload is huge. In this paper, the original 2098 images were manually labelled using the LabelImg labelling software, and then the amplified data samples were automatically labelled using the automatic labelling algorithm. After the labelling was completed, the automatically labelled samples were manually checked and manually adjusted using the LabelImg software to ensure the reliability of the data.
\subsection{Self-made wood panel defect data set}
In this paper, the high-resolution wood surface defect images published by Bodzas et al. were used to obtain annotation files in xml format corresponding to the samples using a combination of manual and automatic methods. To meet the training requirements of the network model, the xml files needed to be converted into txt format files, and then a random data division method was applied to select 90\% of the enhanced data set as the training set and the remaining 10\% as the test set. The expanded number of defect annotations is shown in Table~\ref{tab1}.

\begin{table}[t]
\centering
\caption{Number of dataset annotations.\label{tab1}}
{\scriptsize
\begin{tabular}{lcccc}
\br
Labels                   & Dead Knot & Live Knot & Knot with crack & Crack \\ \mr
Number of pre-expansions & 1348      & 1233      & 848             & 590   \\ 
Number after expansion   & 3139      & 2351      & 2672            & 2817  \\ \br
\end{tabular}
}
\end{table}

\subsection{Experimental environment}
The software algorithm implementation starts with the experimental environment setup. Table~\ref{tab2} shows the description of the hardware environment configuration for this experiment. All experiments in this paper are based on the following environment configuration and the same dataset. The training parameters were set as follows: input image pixel size of 640x640, the total number of training rounds of 250, the initial learning rate of 0.01, cosine annealing rate of 0.1, learning rate momentum of 0.937, weight decay coefficient of 0.0005, batch size of 16, SGD was used as the optimisation function and IoU threshold was set to 0.5 for testing.
\begin{table}[ht]
\centering
\caption{Experimental environment configuration.\label{tab2}}
{\scriptsize
\begin{tabular}{ll}
\br
\multicolumn{1}{l}{Software and Hardware} & \multicolumn{1}{l}{Name Versions}                \\ 
\mr
Operating systems                           & Windows 11 64 bit                                 \\
Central processing   unit                   & AMD Ryzen 7 5800H   with Radeon Graphics 3.20 GHz \\
Graphics   processors                       & GeForce RTX 3060 6GB+CUDA11.6.0                   \\
Programming   Languages                     & Python3.8.13                                      \\
Programming   environment                   & PyCharm 2021                                      \\
Deep learning   frameworks                  & Pytorch1.11.1    \\      
\br
\end{tabular}
}
\end{table}

\subsection{Model evaluation indicators}
The main goal of the improved model in this paper is to make the model as light as possible while maintaining its accuracy. The FPS indicates the frame rate per second, i.e. the number of images processed per second, and is used to evaluate the detection speed of the model. The number of parameters (Params) and the number of floating point operations (FLOPs) correspond to the spatial and temporal complexity of the model, respectively; DK is the accuracy of detecting dead knot defects, LK is the accuracy of detecting live knot defects, KC is the accuracy of detecting nodal pinch defects and CR is the accuracy of detecting crack defects. The evaluation accuracy (AP) is calculated by the accuracy and recall of the model training data. The full class average accuracy (mAP) is the average of the average accuracy AP for each class with the following equation:

\begin{equation}
Precision =\frac{T P}{T P+F P}
\end{equation}

\begin{equation}
Recall=\frac{T P}{T P+F N}
\end{equation}

\begin{equation}
A P=\int_{0}^{1} P d R
\end{equation}

\begin{equation}
m A P=\frac{1}{n} \sum_{i=0}^{n} A P_{i}
\end{equation}
Which: TP is the number of positive samples identified correctly, FP is the number of negative samples identified as positive, FN is the number of positive samples missed; n is the number of categories in the training dataset, i is the current category number.

\section{Results and analysis}
\subsection{comparison experiment with different backbone networks}

\begin{table}[b]
\centering
\caption{Accuracy results of YOLOv5 experiments on different backbone networks.\label{tab3}}
{\scriptsize
\begin{tabular}{llccccc}
\br
\multirow{2}{*}{No.} & \multirow{2}{*}{Backbone} & \multicolumn{4}{c}{AP (\%) \faLongArrowAltUp} & \multirow{2}{*}{mAP (\%) \faLongArrowAltUp} \\ \cline{3-6}
                        &                                   & DK    & LK   & KC   & CR   &           \\
\mr
1                 & C3Net                             & \textbf{91.6}  & \textbf{88.2} & \textbf{97.5} & \textbf{93.3} & \textbf{92.6}                      \\
2               & MobileNetv3                       & 86.5  & 85.9 & 97.2 & 88.1 & 89.4                       \\
3               & ShuffleNetv2                      & 85.2  & 83.4 & 96.7 & 84.6 & 87.4                        \\
4               & ShuffleNetv2\_Stem                & 86.7  & 80.5 & 97.5 & 86.0 & 87.7                            \\
5               & ShuffleNetv2\_focus               & 81.3  & 81.7 & 94.7 & 81.2 & 84.7                           \\
6               & ShuffleNetv2\_Stem\_SPPF          & 87.0  & 82.6 & 97.0 & 87.5 & 88.5                     \\
\br
\end{tabular}
}
\end{table}

In this paper, we use six ways to improve the backbone network of YOLOv5 for comparison experiments and then verify the improvement effect of using YOLOv5-3S as the backbone network. The main experimental procedure: firstly, the original backbone network (i.e., C3Net) of YOLOv5s is replaced with the lightweight convolutional neural network MobileNetv3 and ShuffleNetv2, which are reconstructed and combined to obtain YOLOv5Mv3 and YOLOv5Sv2. Secondly, the backbone network of YOLOv5Sv2 is improved and reconstructed by adding the Stem Block module to reconstruct to become the YOLOv5-2S model, adding the focus module network module to reconfigure it into the YOLOv5-Sf model, and adding the Stem Block and SPPF network modules to reconfigure it into the YOLOv5-3S model. The original YOLOv5s model was compared with the remaining five lightweight models to verify whether the improved models could effectively compress the model computation, reduce the number of parameters and reduce the model inference time. the evaluation metrics of the six network models are shown in Table~\ref{tab3} and Table~\ref{tab4}.

\begin{table}[t]
\centering
\caption{Comprehensive performance of different backbone network experiments with YOLOv5.\label{tab4}}
{\scriptsize
\begin{tabular}{llcccc}
\br
\multirow{2}{*}{No.}  & \multirow{2}{*}{Backbone}  & \multirow{2}{*}{FPS \faLongArrowAltUp} & \multirow{2}{*}{Params (M) \faLongArrowAltDown} & \multirow{2}{*}{FLOPs (G) \faLongArrowAltDown} \\ 
                        &                                      &                          &                      &                            &                           \\
\mr
1                & C3Net                                                  & 177                  & 7.02                       & 16.0                      \\
2                & MobileNetv3                                           & 180                  & 3.37                       & 6.3                       \\
3                & ShuffleNetv2                                           & 195                  & 0.84                       & 2.0                       \\
4                & ShuffleNetv2\_Stem                                    & 190                  & 0.85                       & 2.4                       \\
5                & ShuffleNetv2\_focus                                    & 178                  & 0.85                       & 7.8                       \\
6                & ShuffleNetv2\_Stem\_SPPF                               & \textbf{197}                  & \textbf{1.01}                       & \textbf{2.6} \\
\br
\end{tabular}
}
\end{table}

From the results in Table~\ref{tab3} and Table~\ref{tab4}, it can be seen that all five different backbone network models have a great improvement in lightness compared with the original YOLOv5 backbone model, effectively reducing the model computation, reducing the model size and improving the model detection speed, but at the same time, the detection accuracy also decreases. From the comparison experiments using the lightweight convolutional neural networks MobileNetv3 and ShuffleNetv2 to replace the backbone network in YOLOv5, the average network detection accuracy of the YOLOv5Sv2 model is 2.0 percentage points lower than that of the YOLOv5Mv3 model, but significantly stronger than the latter in terms of model lightweight, and this experimental result validates the use of The experimental results validate the effectiveness of model lightweight using the ShuffleNetv2 network. The reduction in computation and the number of model parameters is easier to deploy for embedded platforms and can reduce the hardware requirements for model training and prediction. To compensate for the loss of accuracy, the ShuffleNetv2 backbone network was improved to compensate for the reduction in accuracy that follows from the lightweight of the backbone network. By replacing the first layer of the original ShuffleNetv2 network with the Stem Block, experimental results show that there is no significant change in the number of network model parameters or computational effort, and the average detection accuracy increases by 0.2 percentage points. However, using the same method, adding the focus module not only increased the computational power of the model but also decreased the detection accuracy. The results show that the Stem Block can improve the extraction ability of the network features. 

Finally, based on the network model of YOLOv5-2S, the detection accuracy of the model was increased from 87.7\% to 88.5\% by adding the Spatial Pyramid Pooling Fast (SPPF) module to the last layer of the backbone network. Although the improved model has a significant increase in the number of parameters and computation, the mAP is increased by 0.8\% and the detection speed does not change significantly; the model balances the accuracy with the number of parameters. 
The results of this comparative experiment show that the improved YOLOv5-3S network, can reduce classification and confidence loss, facilitate the extraction of feature maps, express rich feature information, and provide a significant improvement in accuracy compared with the other four lightweight detection models. The effectiveness of the lightweight nature of the YOLOv5-3S network was verified through experiments, and the use of YOLOv5-3S deployed to an embedded platform for real-time detection has considerable advantages.

\subsection{YOLOv5-LW network method ablation experiment}
After the backbone network is lightly improved, it is far from sufficient to rely solely on the Stem Block and SPPF modules to compensate for the accuracy of the network model. Therefore, based on the YOLOv5-3S model, ECA attention mechanism, and Multi-scale Bi-directional Feature Pyramid Network (MBiFPN) improvements are used to increase the detection of fine target defects on the surface of the boards to further improve the detection performance of the network model.

To verify the detection performance of the algorithm proposed in this paper, ablation experiments were carried out on the proposed improved method, using a home-made wooden board defect dataset as the training object, using the same training techniques as well as hyperparameter settings for each group of experiments, adding the ECA attention mechanism, and Multi-scale Bi-directional Feature Pyramid Network (MBiFPN), improvement methods in turn, and the evaluation index results of the experiments are shown in Table~\ref{tab5}.

\begin{table}[b]
\centering
\caption{YOLOv5-LW network method ablation experiments.\label{tab5}}
{\scriptsize
\begin{tabular}{ccccc}
\br
No. & ECA & MBiFPN  & mAP (\%) \faLongArrowAltUp & FPS \faLongArrowAltUp \\
\mr
1     &     &              & 88.5    & \textbf{197}      \\
2     & \checkmark   &              & 89.3    & 196      \\
3     & \checkmark   & \checkmark          & \textbf{92.8}    & 195   \\
\br
\end{tabular}
}
\end{table}

As seen in Table~\ref{tab5}, the results of comparing No.1 with No.2 indicated that by adding the Efficient Channel Attention (ECA) mechanism to the backbone network only, the mAP improved from 88.5\% to 89.3\% with the same number of model parameters and computational effort, indicating that the improved amplification increased the ability of the network model to focus on critical information and enhanced the network's ability to detect fine defects. To further improve the model's detection of fine defects, the mAP was improved to 92.8\% by building on No.2, which used MBiFPN in the feature fusion network section to reduce feature loss. 
From the results of the ablation experiments, the detection of fine defects can be increased by the ECA attention mechanism and MBiPFN feature fusion, and although there is an increase in the number of parameters and computational effort in the network model, the model performance is improved and the detection speed is not greatly reduced. Compared to the original model YOLOv5s, the improved model has reduced the number of parameters by 27.78\%, compressed the computation by 41.25\% and improved the detection inference speed by 10.16\%. Compared to No.1, the improvement effect has improved in the detection of fine defects. The experiments in this paper validate the performance of the improved model in this paper. Based on YOLOv5-3S, although the number of parameters and the computational effort of the model has increased, the detection accuracy of the model has increased and the inference time has not decreased significantly. The increase in computation is well worth it.
\begin{figure}[t]
\centering
\includegraphics[width=1\textwidth]{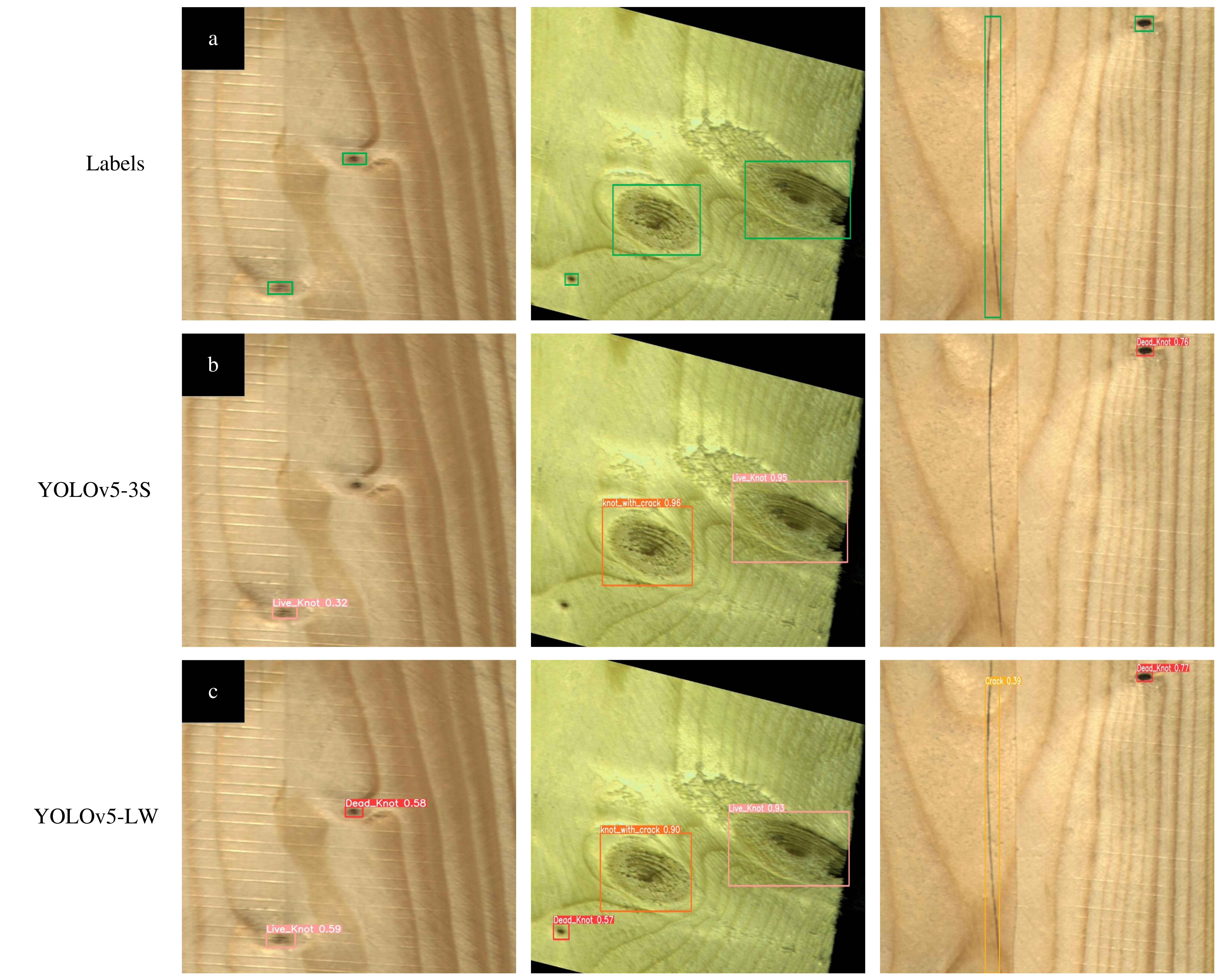}
\caption{Comparison of the detection results of the YOLOv5-LW network method: (a) shows the model input label, (b) YOLOv5-3S model test results, and (c) YOLOv5-LW method test results.\label{fig6}}
\end{figure}

Figure~\ref{fig6} shows a comparison of the detection effect of the YOLOv5-LW method ablation experiment. From figure (b), it can be seen that there are 3 minor defects with missed detection, as long as they are 2 dead knot defects and 1 minor crack defect, which indicates that the model has some difficulties in detecting minor defects, and there are only 2 places in the figure where the confidence level exceeds 0.9, and the overall identification confidence level is low. From (b) compared with (c), 
the fine defects missed in (b) are easier to detect, verifying that the improved method can improve the model's ability to sense fine defects, in addition to the confidence of the defect targets are improved, indicating that the model performance can be improved by using the attention mechanism, weighted feature fusion network. Through ablation experiments, it is verified that the network model proposed in this paper makes full use of the target information of the feature map to improve the model detection accuracy. Therefore, the network model proposed in this paper can be deployed to embedded devices for real-time detection.

\subsection{Comparative Experiments with classical detection networks}
\begin{figure}[t]
\centering
\includegraphics[width=1\textwidth]{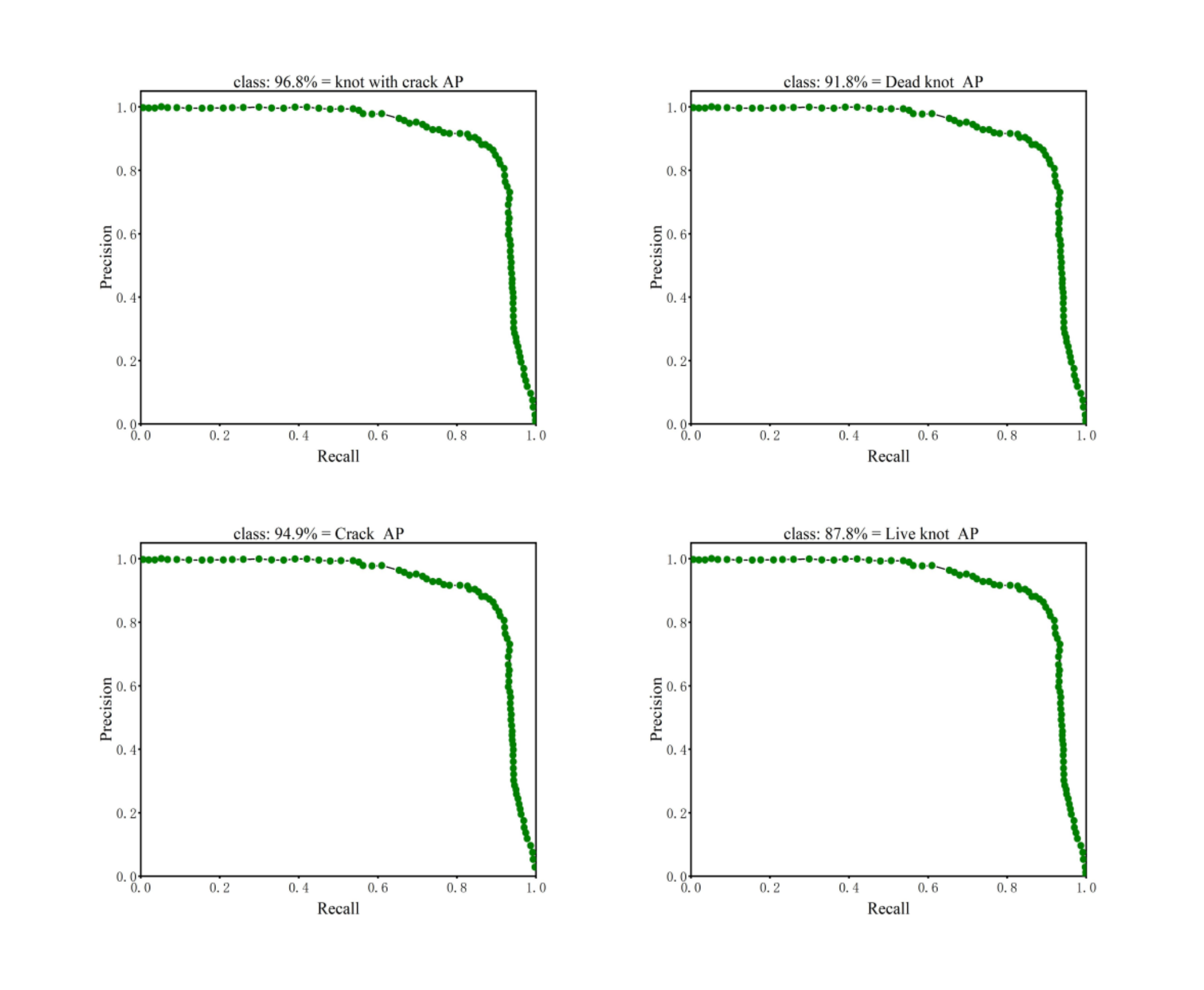}
\caption{YOLOv5-LW method PR graph: contains 4 types of defects: crack, dead knot, live knot, and knot with crack, it can be seen that there is a high recognition class and recall rate for fine defects.\label{fig7}}
\end{figure}

To further validate the effectiveness of the YOLOv5-LW method proposed in this paper, network models with different target detection algorithms were trained on a homemade plank dataset. To ensure the reliability of the experimental data, the same hyperparameters were used to train the YOLOv5s \cite{wang2021channel}, YOLOv4 \cite{Song_2023}, YOLOv3 \cite{Liu_2020}, SSD \cite{Shan_2019} and Faster RCNN \cite{Wu_2019} networks. Figure~\ref{fig7} shows the PR (Precision Recall) curve for the YOLOv5-LW method, and the evaluation index results are shown in Table~\ref{tab6}.

\begin{table}[b]
\caption{Comparison experiments of classical detection networks. A \textcolor{gray}{grey} row indicates a larger backbone that does not allow for a fair comparison of accuracy advantages.\label{tab6}}
\centering
{\tiny
\begin{tabular}{llcccccccc}
\br
\multicolumn{1}{l}{\multirow{2}{*}{Model}} & \multirow{2}{*}{Backbone} & \multicolumn{4}{c}{AP (\%) \faLongArrowAltUp} & \multirow{2}{*}{mAP (\%) \faLongArrowAltUp} & \multirow{2}{*}{FPS \faLongArrowAltUp} & \multirow{2}{*}{Params (M) \faLongArrowAltDown} & \multirow{2}{*}{FLOPs (G) \faLongArrowAltDown} \\ \cline{3-6}
\multicolumn{1}{l}{}         &               & DK    & LK    & KC   & CR   &                          &                      &                            &                           \\
\mr
Faster Rcnn \cite{Wu_2019}        & Resnet50                          & 65.9  & 78.7  & 97.4 & 83.5 & 81.4                    & 11                   & 44.10                      & 91.0                      \\
YOLOv4 \cite{Song_2023}          & CSPDarknet53                             & 69.9  & 44.3  & 82.6 & 58.5 & 63.8                     & 36                   & 63.95                      & 59.9                      \\
YOLOv3 \cite{Liu_2020}     & Darket53                                 & 88.4  & 86.4  & 97.5 & 94.2 & 91.6                    & 55                   & 61.51                      & 154.6                     \\
SSD \cite{Shan_2019}    & vgg                                   & 76.4  & 77.3  & 63.3 & 78.0 & 73.7                    & 75                   & 24.01                      & 61.05                    \\
YOLOv3-tiny \cite{Liu_2020}     & Darknet53-tiny                                  & 69.9  & 64.1  & 87.8 & 65.6 & 71.8                     & 105                   & 8.67                      & 12.9                     \\
\textcolor{gray}{YOLOv5s \cite{wang2021channel}}      & \textcolor{gray}{C3Net}                                & \textcolor{gray}{91.6}  & \textcolor{gray}{\textbf{88.2}}  & \textcolor{gray}{\textbf{97.5}} & \textcolor{gray}{93.3} & \textcolor{gray}{92.6}                     & \textcolor{gray}{177}                  & \textcolor{gray}{7.02}                       & \textcolor{gray}{16.0}                      \\
\textbf{YOLOv5-LW(Ours)}     & \textbf{$S^3$Net}                                 & \textbf{91.8}  & 87.8  & 96.8 & \textbf{94.6} & \textbf{92.8}                     & \textbf{195}                  & \textbf{5.07}                        & \textbf{9.4}    \\
\br
\end{tabular}
}
\end{table}

The experimental results in Table~\ref{tab6} show that the two-stage target detection algorithm Faster Rcnn \cite{Wu_2019}, compared to the single-stage target detection algorithm YOLOv4 \cite{Song_2023} with a larger number of parameters and computation, has a much higher detection accuracy, but the YOLO series has a great improvement in detection speed and a number of parameters, so it is more utilised for network deployment to embedded devices for real-time detection of wood panel defects. The YOLOv5-LW method proposed in this paper outperforms the other five models, which, in comparison, have weaker generalisation capabilities, slower detection speed, and a large number of parameters that are not easily deployed. Therefore, the choice of YOLOv5-LW as the wood panel defect detection model has better convincing power. As can be seen from  Figure~\ref{fig7}, the two metrics of accuracy and recall interact with each other, with the PR curve better reflecting the improved model having a higher recognition rate for fine defects.  

As can be seen from Table~\ref{tab6}, the improved YOLOv5-LW model is not much better than the original model before improvement in terms of accuracy, mainly due to the fact that the original backbone network (i.e., C3Net) is larger, resulting in a more adequate backbone extraction feature map, while the improved S3Net backbone network is smaller, resulting in a detection effect that is obviously not better than the original network (i.e., C3Net) in terms of accuracy, but this paper's improved method improves accuracy by 4.3\% on a lightweight basis and reduces the overall parameter volume by 27.78\%, compressed the computational volume by 41.25\%, improved the detection inference speed by 10.16\% and improved the detection accuracy for two types of fine defects, dead knots and cracks, by 0.2\% and 1.3\% respectively, compared to the original model before the improvement. The experimental results show that the loss of accuracy due to lightweight can be compensated by the Stem Block and SPPF modules as well as the attention mechanism and bi-directional feature fusion network. The improved network model has enhanced generalisation capability to targets with fine defects, and the network has lower hardware requirements, is easier to deploy to embedded platforms, and can be applied to tasks with higher defect localisation requirements, meeting the requirements for real-time detection of wood panel defects.

\section{Conclusions}
In this paper, we design a wood panel fine defect detection algorithm on the basis of YOLOv5 network, firstly, we adopt the lightweight network ShuffleNetv2 for backbone lightweight, which greatly reduces the number of model parameters and computation, and improves the detection inference time of the model; secondly, we reconfigure the backbone network through Stem Block and Spatial Pyramid Pooling Fast (SPPF) modules to improve the model detection performance, followed by combining Efficient Channel Attention (ECA) mechanism to network's ability to focus on key information is improved; and the MBiFPN bi-directional feature fusion network is introduced to reduce feature loss, enrich local and detailed features, improve the detection capability of fine defects; finally, experimental validation of the algorithm proposed in this paper on a homemade dataset shows that the accuracy of the improved model reaches 92.8\%, the number of parameters is reduced by 27.78\%, the computational volume is compressed by 41.25\%, the detection inference speed is improved by 10.16\%, and the detection accuracy of two types of small defects, dead knots and cracks, is also improved by 0.2\% and 1.3\% respectively. Compared to other classical models, the algorithm proposed in this paper is more conducive to the deployment of embedded devices and has better performance in terms of detection speed and detection of fine targets. However, there is still room for further optimisation of the model to improve its accuracy.

\textbf{Future Work.} the algorithms proposed in this paper will be deployed to embedded devices for field applications to further investigate the practicality of the devices.



\maketitle

\end{document}